\def\year{2016}\relax
\documentclass[letterpaper]{article}
\usepackage{aaai16}
\usepackage{times}
\usepackage{helvet}
\usepackage{courier}
\usepackage{colortbl}
\usepackage{graphicx}
\title{How Important Is Weight Symmetry in Backpropagation?}
\author{Qianli Liao \and Joel Z. Leibo \and Tomaso Poggio\\
Center for Brains, Minds and Machines, McGovern Institute \\
Massachusetts Institute of Technology \\
77 Massachusetts Ave., Cambridge, MA, 02139, USA
}

\begin{document}       
\maketitle
\begin{abstract}
Gradient backpropagation (BP) requires symmetric feedforward and feedback connections---the same weights must be used for forward and backward passes. This ``weight transport problem'' (Grossberg 1987) is thought to be one of the main reasons to doubt BP's biologically plausibility. Using 15 different classification datasets, we systematically investigate to what extent BP really depends on weight symmetry. In a study that turned out to be surprisingly similar in spirit to Lillicrap et al.'s demonstration (Lillicrap et al. 2014) but orthogonal in its results, our experiments indicate that: (1) the magnitudes of feedback weights do not matter to performance (2) the signs of feedback weights do matter---the more concordant signs between feedforward and their corresponding feedback connections, the better (3) with feedback weights having random magnitudes and 100\% concordant signs, we were able to achieve the same or even better performance than SGD. (4) some normalizations/stabilizations are indispensable for such asymmetric BP to work, namely Batch Normalization (BN) (Ioffe and Szegedy 2015) and/or a ``Batch Manhattan'' (BM) update rule.
\end{abstract}

\newcommand{\specialcell}[2][c]{%
  \begin{tabular}[#1]{@{}c@{}}#2\end{tabular}}

\section{Introduction}
Deep Neural Networks (DNNs) have achieved remarkable performance in many domains \cite{Krizhevsky2012,Abdel-Hamid2012,hinton2012deep,mikolov2013distributed,taigman2014deepface,graves2014neural}. The simple gradient backpropagation (BP) algorithm has been the essential ``learning engine'' powering most of this work. 

Deep neural networks are universal function approximators \cite{hornik1989multilayer}. Thus it is not surprising that solutions to real-world problems exist within their configuration space. Rather, the real surprise is that such configurations can actually be discovered by gradient backpropagation.

The human brain may also be some form of DNN. Since BP is the most effective known method of adapting DNN parameters to large datasets, it becomes a priority to answer: could the brain somehow be implementing BP? Or some approximation to it? 

For most of the past three decades since the invention of BP, it was generally believed that it could not be implemented by the brain   \cite{crick1989recent,mazzoni1991more,o1996biologically,chinta2012adaptive,bengio2015towards}.  BP seems to have three biologically implausible requirements: (1) feedback weights must be the same as feedforward weights (2) forward and backward passes require different computations, and (3) error gradients must somehow be stored separately from activations. 

One biologically plausible way to satisfy requirements (2) and (3) is to posit a distinct ``error network'' with the same topology as the main (forward) network but used only for backpropagation of error signals. The main problem with such a model is that it makes requirement (1) implausible. There is no known biological way for the error network to know precisely the weights of the original network. This is known as the ``weight transport problem'' \cite{grossberg1987competitive}. In this work we call it the ``weight symmetry problem''. It is arguably the crux of BP's biological implausibility. 

In this report, we systematically relax BP's weight symmetry requirement by manipulating the feedback weights. We find that some natural and biologically plausible schemes along these lines lead to exploding or vanishing gradients and render learning impossible. However, useful learning is restored if a simple and indeed \emph{more} biologically plausible rule called Batch Manhattan (BM) is used to compute the weight updates. Another technique, called Batch Normalization (BN) \cite{ioffe2015batch}, is also shown effective. When combined together, these two techniques seem complementary and significantly improve the performance of our asymmetric version of backpropagation.

The results are somewhat surprising: if the aforementioned BM and/or BN operations are applied, the magnitudes of feedback weights turn out not to be important.  A much-relaxed \emph{sign-concordance} property is all that is needed to attain comparable performance to mini-batch SGD on a large number of tasks.

Furthermore, we tried going beyond sign concordant feedback. We systematically reduced the probability of feedforward and feedback weights having the same sign (the \emph{sign concordance probability}). We found that the effectiveness of backpropagation is strongly dependent on high sign concordance probability. That said, completely random and fixed feedback still outperforms chance e.g., as in the recent work of Lillicrap et al. \cite{lillicrap2014random}.

Our results demonstrate that the perfect forward-backward weight symmetry requirement of backpropagation can be significantly relaxed and strong performance can still be achieved.  To summarize, we have the following conclusions: \\
\textbf{(I)} \textbf{The magnitudes of feedback weights do not matter to performance}. This surprising result suggests that our theoretical understanding of why backpropagation works may be far from complete.\\
\textbf{(II)} Magnitudes of the weight updates also do not matter.\\
\textbf{(III)} Normalization / stabilization methods such as Batch Normalization and Batch Manhattan are necessary for these asymmetric backpropagation algorithms to work. Note that this result was missed by previous work on random feedback weights \cite{lillicrap2014random}.  \\
\textbf{(IV)} Asymmetric backpropagation algorithms evade the weight transport problem. Thus it is plausible that the brain could implement them. \\
\textbf{(V)} These results indicate that sign-concordance is very important for achieving strong performance. However, even fixed random feedback weights with Batch Normalization significantly outperforms chance. This is intriguing and motivates further research. \\
\textbf{(VI)} Additionally, we find Batch Manhattan to be a very simple but useful technique in general. When used with Batch Normalization, it often improves the performance. This is especially true for smaller training sets.

\section{Asymmetric Backpropagations}
\label{sec:abp}

\begin{figure}[h]
\centering
 \includegraphics{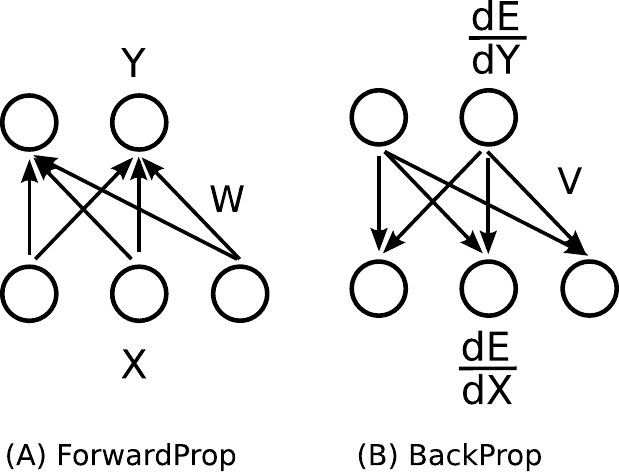}
  \caption{A simple illustration of backpropagation}
\label{fig:bp_basics}
\end{figure}

A schematic representation of backpropagation is shown in Fig. \ref{fig:bp_basics}. Let $E$ be the objective function.  Let $W$ and $V$ denote the feedforward and feedback weight matrices respectively.  Let $X$ denote the inputs and $Y$ the outputs. $W_{ij}$ and $V_{ij}$ are the feedforward and feedback connections between the $j$-th output $Y_j$ and the $i$-th input $X_i$, respectively. $f(.)$ and $f'(.)$ are the transfer function and its derivative. Let the derivative of the $i$-th input with respect to the objective function be $\frac{\partial E}{\partial X_i}$, the formulations of forward and back propagation are as follows:  \\
%Forward propagation:
\begin{equation}
Y_j = f(N_j), where\,\, N_j = \sum_i W_{ij}X_i 
\end{equation}
%Back propagation:
\begin{equation}
\frac{\partial E}{\partial X_i} = \sum_j V_{ij}f'(N_j)\frac{\partial E}{\partial Y_j}
\end{equation}

The standard BP algorithm requires $V=W$.  We call that case \emph{symmetric backpropagation}. In this work we systematically explore the case of \emph{asymmetric backpropagation} where $V \neq W$.

By varying $V$, one can test various asymmetric BPs.  Let $sign()$ denote the function that takes the sign (-1 or 1) of each element. Let $\circ$ indicate element-wise multiplication. $M,S$ are matrices of the same size as $W$. $M$ is a matrix of uniform random numbers $\in [0,1]$ and $S_p$ is a matrix where each element is either $1$ with probability $1-p$ or $-1$ with probability $p$. We explored the following choices of feedback weights $V$ in this paper: \\
1. \textbf{Uniform Sign-concordant Feedbacks (uSF)}:\\ $V = sign(W)$ \\
2. \textbf{Batchwise Random Magnitude Sign-concordant Feedbacks (brSF)}:\\ $V = M \circ sign(W)$, where $M$ is redrawn after each update of $W$ (i.e., each mini-batch). \\
3. \textbf{Fixed Random Magnitude Sign-concordant Feedbacks (frSF)}:\\ $V = M \circ sign(W)$, where $M$ is initialized once and fixed throughout each experiment.  \\
4. \textbf{Batchwise Random Magnitude p-percent-sign-concordant Feedbacks (brSF-p)}:\\ $V = M  \circ  sign(W) \circ S_p$, where $M$ and $S_p$ is redrawn after each update of $W$  (i.e., each mini-batch). \\
5. \textbf{Fixed Random Magnitude  p-percent-sign-concordant Feedbacks (frSF-p)}:\\ $V = M  \circ  sign(W) \circ S_p$, where $M$ and $S_p$ is initialized once and fixed throughout each experiment.  \\
6. \textbf{Fixed Random Feedbacks (RndF)}:\\  Each feedback weight is drawn from a zero-mean gaussian distribution and fixed throughout each experiment: $V \sim \mathcal{N} (0,\sigma^2)$, where $\sigma$ was chosen to be 0.05 in all experiments. \\

The results are summarized in the Section \ref{sec:exp}. The performances of 1, 2 and 3, which we call \textbf{strict sign-concordance} cases, are shown in Experiment A. The performances of 4 and 5 with different choices of $p$, which we call \textbf{partial sign-concordance} cases, are shown in Experiment B. The performances and control experiments about setting 6, which we call \textbf{no concordance} cases, are shown in Experiments C1 and C2.

\section{Normalizations/stabilizations are necessary for ``asymmetric'' backpropagations}
\label{sec:normalizations}
\textbf{Batch Normalization (BN)} \\
Batch Normalization (BN) is a recent technique proposed by \cite{ioffe2015batch} to reduce ``internal covariate shift'' \cite{ioffe2015batch}. The technique consists of element-wise normalization to zero mean and unit standard deviation. Means and standard deviations are separately computed for each batch. Note that in \cite{ioffe2015batch}, the authors proposed the use of additional learnable parameters after the whitening. We found the effect of this operation to be negligible in most cases. Except for the ``BN'' and ``BN+BM'' entries (e.g., in Table \ref{tab:main}), we did not use the learnable parameters of BN. Note that batch normalization may be related to the homeostatic plasticity mechanisms (e.g., Synaptic Scaling) in the brain \cite{Turrigiano2004,stellwagen2006synaptic,turrigiano2008self}.

\textbf{Batch Manhattan (BM)} \\
We were first motivated by looking at how BP could tolerate noisy operations that could be seen as more easily implementable by the brain. We tried relaxing the weight updates by discarding the magnitudes of the gradients. Let the weight at time $t$ be $w(t)$, the update rule is:

\begin{equation}
  w(t+1) = w(t) + \eta*\tau(t)
\end{equation}
where $\eta$ is the learning rate.

We tested several settings of $\tau(t)$ as follows: \\
\textbf{Setting 0 (SGD)}: $\tau(t) = -\sum_b \frac{\partial E}{\partial w} +  m*\tau(t-1) - d*w(t)$ \\
\textbf{Setting 1}: $\tau(t) = -sign(\sum_b \frac{\partial E}{\partial w}) +  m*\tau(t-1) - d*w(t)$ \\
\textbf{Setting 2}: $\tau(t) =  sign(-sign(\sum_b \frac{\partial E}{\partial w}) + m*\tau(t-1) - d*w(t))$ \\
\textbf{Setting 3}: $\tau(t) =  sign(\kappa(t))$ \\ where $\kappa(t) =  -sign(\sum_b \frac{\partial E}{\partial w}) + m*\kappa(t-1) - d*w(t)$ \\
\\
where $m$ and $d$ are momentum and weight decay rates respectively. $sign()$ means taking the sign (-1 or 1), $E$ is the objective function, and $b$ denotes the indices of samples in the mini-batch. Setting 0 is the SGD algorithm (by ``SGD'' in this paper, we always refer to the mini-batch version with momentum and weight decay). Setting 1 is same as 0 but rounding the accumulated gradients in a batch to its sign. Setting 2 takes an extra final sign after adding the gradient term with momentum and weight decay terms. Setting 3 is something in between 1 and 2, where an final sign is taken, but not accumulated in the momentum term. 

We found these techniques to be surprisingly powerful in the sense that they did not lower performance in most cases (as long as learning rates were reasonable). In fact, sometimes they improved performance. This was especially true for smaller training sets. Recall that asymmetric BPs tend to have exploding/vanishing gradients, these techniques are immune to such problems since the magnitudes of gradients are discarded.

We also found that the performance of this technique was influenced by batch size on some experiments. In the cases of very small batch sizes, discarding the magnitudes of the weight updates was sometimes detrimental to performance.

This class of update rule is very similar to a technique called the Manhattan update rule, which can be considered as a simplified version of Rprop \cite{riedmiller1993direct}. We suggest calling it ``Batch Manhattan'' (BM) to distinguish it from the stochastic version \cite{zamanidoostmanhattan}. By default, we used setting 1 for BM throughout the Experiments A, B, C1 and C2. The ``miscellaneous experiment'' at the end of the Section \ref{sec:results} demonstrates that settings 1, 2 and 3 give similar performances, so the conclusions we draw broadly apply to all of them.

\section{Related Work}

Since the invention of backpropagation (BP) \cite{rumelhart1988learning}, its biological plausibility has been a long-standing controversy.  Several authors have argued that   BP is not biologically plausible \cite{crick1989recent,mazzoni1991more,o1996biologically,chinta2012adaptive,bengio2015towards}. Various biologically plausible modifications have been proposed. Most involve bidirectional connections e.g. Restricted Boltzmann Machines \cite{hinton2006reducing,smolensky1986information} and so-called recirculation algorithms \cite{hinton1988learning,o1996biologically} which despite their name provided,  in the case of an autoencoder, an elegant early demonstration that adaptive backwards weight can work without being identical to the forward ones.  Recently, there have also been BP-free auto-encoders \cite{bengio2014auto} based on ``target propagation'' \cite{le1986learning}.

The most relevant work to ours is a recent paper by Lillicrap et al. \cite{lillicrap2014random} of which we became aware after most of this work was done. Lillicrap et al. showed that fixed random feedback weights can support the learning of good representations for several simple tasks: (i) approximating a linear function, (ii) digit recognition on MNIST and (iii) approximating the outputs of a random 3 or 4 layer nonlinear neural network. Our work is very similar in spirit but rather different and perhaps complementary in its results, since we conclude that signs must be concordant between feedforward and corresponding feedback connections for consistent good performance, whereas the magnitudes do not matter, unlike Lilicrap et al. experiments in which both signs and magnitudes were random (but fixed). To explain the difference in our conclusions, it is useful to consider the following points: \\
\textbf{1.}  We systematically explored performance of the algorithms using 15 different datasets because simple tasks like MNIST by themselves do not always reveal differences between algorithms. \\
\textbf{2.} We tested deeper networks, since the accuracy of asymmetric BP's credit assignment may critically attenuate with depth (for task (i) and (ii) Lillicrap et al. used a 3-layer (1 hidden layer) fully-connected network, and for task (iii) they used a 3 or 4 layer fully-connected network, whereas in most of our experiments, we use deeper and larger CNNs as shown in Table \ref{tab:arch}).\\
\textbf{3.}  We found that local normalizations/stabilizations is critical for making  asymmetric BP algorithms work. As shown by our results in Table \ref{tab:exp_C_and_D}, the random feedbacks scheme (i.e. the ``RndF'' column) suggested by Lillicrap et al. seem to work well only on one or two tasks, performing close to chance on most of them. Only when combined with Batch Normalization (``RndF+BN'' or ``RndF+BN+BM'' in Table \ref{tab:exp_C_and_D}), it appears to become competitive.\\
\textbf{4.}  We investigated several variants of asymmetric BPs such as sign-concordance (Table \ref{tab:main} and \ref{tab:percent_diff}), batchwise-random vs. fixed-random feedbacks (Table \ref{tab:percent_diff}) and learning with clamped layers (Table \ref{tab:exp_C_and_D} Exp. C2).

\definecolor{dG}{rgb}{0.08,0.08,0.7}
\definecolor{dR}{rgb}{0.66,0.66,0}
\newcommand{\tdG}[1]{\textcolor{dG}{\textbf{#1}}}
\newcommand{\tdR}[1]{\textcolor{dR}{\textbf{#1}}}
\newcommand{\mc}[2]{\multicolumn{#1}{c}{#2}}
\definecolor{Gray}{gray}{0.85}
\definecolor{LightCyan}{rgb}{0.85,0.97,0.97}

\begin {table*}
\centering
%\begin{center}
\begin{tabular}{  |c|c|c|c|c|c| }
 \hline
            & All others &        MNIST      &   CIFAR10\&100 & SVHN            & TIMIT-80  \\
 \hline
 \hline
 InputSize  & 119x119x3                 & 28x28x1           & 32x32x3        & 32x32x3         & 1845x1x1 \\
 \hline
 1          & Conv 9x9x48/2             & Conv 5x5x20/1     & Conv 5x5x32/1  & Conv 5x5x20/1   & FC 512 \\
\rowcolor{LightCyan}
 2          & Max-Pool 2x2/2            & Max-Pool 2x2/2    & Max-Pool 3x3/2 & Max-Pool 2x2/2  & FC 256 \\
 3          & Conv 5x5x128/1            & Conv 5x5x50/1     & Conv 5x5x64/1  & Conv 7x7x512/1  & FC 80 \\
\rowcolor{LightCyan}
 4          & Max-Pool 2x2/2            & Max-Pool 2x2/2    & Avg-Pool 3x3/2 & Max-Pool 2x2/2  & \\
 5          & FC max(256,\#Class*3)    & FC 500            & Conv 5x5x64/1  & FC 40           &        \\
\rowcolor{LightCyan}
 6          & FC \#Class*2             & FC 10             & Avg-Pool 3x3/2 & FC 10           & \\
 7          & FC \#Class               &                   & FC 128         &                 &   \\
\rowcolor{LightCyan}
 8          &                           &                   & FC 10/100      &                 & \\
 \hline
\end{tabular}
%\end{center}
\caption {Network architectures used in the experiments: AxBxC/D means C feature maps of size AxB, with stride D. The CIFAR10\&100 architecture has a 2 units zero-padding for every convolution layer and 1 unit right-bottom zero-padding for every pooling layer. The other models do not have paddings. ``FC X'' denotes Fully Connected layer of X feature maps. In the first model, the number of hidden units in FC layers are chosen according to the number of classes (denoted by ``\#Class'') in the classification task. `` max(256,\#Class*3)'' denotes 256 or \#Class*3, whichever is larger. Rectified linear units (ReLU) are used as nonlinearities for all models.  } \label{tab:arch}
\end{table*}

\section{Experiments}
\label{sec:exp}

\subsection{Method}
We were interested in relative differences between algorithms, not absolute performance. Thus we used common values for most parameters across all datasets to facilitate comparison. Key to our approach was the development of software allowing us to easily evaluate the ``cartesian product'' of models (experimental conditions) and datasets (tasks). Each experiment was a \{model,dataset\} pair, which was run 5 times using different learning rates (reporting the best performance). We used MatConvNet \cite{vedaldi15matconvnet} to implement our models.

\subsection{Datasets}
We extensively test our algorithms on 15 datasets of 5 Categories as described below. No data augmentation (e.g., cropping, flip, etc.) is used in any of the experiments. \\
\textbf{Machine learning tasks:} MNIST \cite{mnistWebsite}, CIFAR-10 \cite{krizhevsky2009learning}, CIFAR-100 \cite{krizhevsky2009learning}, SVHN\cite{netzer2011reading}, STL10 \cite{coates2011analysis}. Standard training and testing splits were used. \\
\textbf{Basic-level categorization tasks:} Caltech101 \cite{fei2007learning}: 102 classes, 30 training and 10 testing samples per class. Caltech256-101 \cite{griffin2007caltech}: we train/test on a subset of randomly sampled 102 classes. 30 training and 10 testing per class. iCubWorld dataset \cite{fanello2013icub}: We followed the standard categorization protocol of this dataset.\\
\textbf{Fine-grained recognition tasks:} Flowers17 \cite{nilsback2006visual}, Flowers102 \cite{nilsback2008automated}. Standard training and testing splits were used.  \\
\textbf{Face Identification:} Pubfig83-ID \cite{pinto2011scaling},  SUFR-W-ID \cite{leibo2014subtasks}, LFW-ID \cite{Huang2008} We did not follow the usual (verification) protocol of these datasets. Instead, we performed a 80-way face identification task on each dataset, where the 80 identities (IDs) were randomly sampled. Pubfig83: 85 training and 15 testing samples per ID.  SUFR-W: 10 training and 5 testing per ID. LFW: 10 training and 5 testing per ID. \\
\textbf{Scene recognition:} MIT-indoor67 \cite{quattoni2009recognizing}:  67 classes, 80 training and 20 testing per class \\
\textbf{Non-visual task:} TIMIT-80 \cite{TIMITdatabase}: Phoneme recognition using a fully-connected network. There are 80 classes, 400 training and 100 testing samples per class.

\begin{table*}
\centering
%\begin{center}
\begin{tabular}{|c|ccccccccccc|}
\hline
Experiment A & \small SGD &\small BM &\small BN &\small BN+BM &\small uSF &\small NuSF &\small \specialcell{uSF\\+BM} &\small \specialcell{uSF\\+BN} &\small \specialcell{uSF\\+BN\\+BM} &\small \specialcell{brSF\\+BN\\+BM} &\small \specialcell{frSF\\+BN\\+BM}\\
\hline
MNIST &0.67 &0.99 &0.52 &0.73 &\tdR{88.65} &0.60 &0.95 &0.55 &0.83 &0.80 &0.91\\
\rowcolor{LightCyan}
CIFAR &22.73 &23.98 &\tdG{16.75} &\tdG{17.94} &\tdR{90.00} &\tdR{40.60} &\tdR{26.25} &\tdG{19.48} &\tdG{19.29} &\tdG{18.44} &\tdG{19.02}\\
CIFAR100 &55.15 &\tdR{58.44} &\tdG{49.44} &\tdG{51.45} &\tdR{99.00} &\tdR{71.51} &\tdR{65.28} &57.19 &53.12 &\tdG{50.74} &52.25\\
\rowcolor{LightCyan}
SVHN &9.06 &10.77 &7.50 &9.88 &\tdR{80.41} &\tdR{14.55} &9.78 &8.73 &9.67 &9.95 &10.16\\
STL10 &48.01 &\tdG{44.14} &45.19 &\tdG{43.19} &\tdR{90.00} &\tdR{56.53} &46.41 &48.49 &\tdG{41.55} &\tdG{42.74} &\tdG{42.68}\\
\rowcolor{LightCyan}
Cal101 &74.08 &\tdG{66.70} &\tdG{66.07} &\tdG{61.75} &\tdR{98.95} &\tdG{70.50} &75.24 &\tdG{63.33} &\tdG{60.70} &\tdG{59.54} &\tdG{60.27}\\
Cal256-101 &87.06 &\tdG{83.43} &\tdG{82.94} &\tdG{81.96} &\tdR{99.02} &85.98 &86.37 &\tdG{82.16} &\tdG{80.78} &\tdG{78.92} &\tdG{80.59}\\
\rowcolor{LightCyan}
iCub &57.62 &55.57 &\tdG{46.43} &\tdG{37.08} &\tdR{89.96} &\tdR{66.57} &\tdR{70.61} &\tdR{61.37} &\tdG{48.38} &\tdG{47.33} &\tdG{46.08}\\
Flowers17 &35.29 &\tdG{31.76} &36.76 &32.35 &\tdR{94.12} &\tdR{42.65} &38.24 &35.29 &32.65 &\tdG{29.41} &\tdG{31.47}\\
\rowcolor{LightCyan}
Flowers102 &77.30 &77.57 &75.78 &74.92 &\tdR{99.67} &77.92 &79.25 &\tdG{71.74} &\tdG{73.20} &\tdG{73.31} &\tdG{73.57}\\
PubFig83-ID &63.25 &\tdG{54.42} &\tdG{51.08} &\tdG{41.33} &\tdR{98.75} &\tdR{78.58} &65.83 &\tdG{54.58} &\tdG{40.67} &\tdG{42.67} &\tdG{40.33}\\
\rowcolor{LightCyan}
SUFR-W-ID &80.00 &\tdG{74.25} &\tdG{75.00} &\tdG{65.00} &\tdR{98.75} &\tdR{83.50} &79.50 &\tdG{72.00} &\tdG{65.75} &\tdG{66.25} &\tdG{66.50}\\
LFW-ID &79.25 &\tdG{74.25} &\tdG{73.75} &\tdG{55.75} &\tdR{98.75} &\tdR{85.75} &80.75 &\tdG{73.75} &\tdG{56.25} &\tdG{57.25} &\tdG{55.75}\\
\rowcolor{LightCyan}
Scene67 &87.16 &85.37 &86.04 &\tdG{82.46} &\tdR{98.51} &88.21 &87.09 &87.09 &\tdG{81.87} &\tdG{82.31} &\tdG{81.79}\\
TIMIT80 &23.04 &25.92 &23.92 &24.40 &23.60 &\tdR{29.28} &25.84 &25.04 &25.12 &25.24 &24.92\\
\hline
\end{tabular}
%\end{center}
\caption{\textbf{Experiment A:} The magnitudes of feedbacks do not matter. Sign concordant feedbacks can produce strong performance. Numbers are error rates (\%). \tdR{Yellow}: performances worse than baseline(SGD) by 3\% or more.  \tdG{Blue}: performances better than baseline(SGD) by 3\% or more.  } \label{tab:main}
\end{table*}

\subsection{Training Details}
 The network architectures for various experiments are listed in Table \ref{tab:arch}. The input sizes of networks are shown in the second row of the table. All images are resized to fit the network if necessary.

Momentum was used with hyperparameter 0.9 (a conventional setting). All experiments were run for 65 epochs. The base learning rate: 1 to 50 epochs $5*10^{-4}$, 51 to 60 epochs $5*10^{-5}$, and 61 to 65 epochs $5*10^{-6}$. All models were run 5 times on each dataset with base learning rate multiplied by 100, 10, 1, 0.1, 0.01 respectively. This is because different learning algorithms favor different magnitudes of learning rates. The best validation error among all epochs of 5 runs was recorded as each model's final performance. The batch sizes were all set to 100 unless stated otherwise. All experiments used a softmax for classification and the cross-entropy loss function. For testing with batch normalization, we compute exponential moving averages (alpha=0.05) of training means and standard deviations over 20 mini batches after each training epoch.

\subsection{Results} 
\label{sec:results} 

\subsubsection{Experiment A: sign-concordant Feedback} 

In this experiment, we show the performances of setting 1, 2 and 3 in Section \ref{sec:abp}, which we call \textbf{strict sign-concordance} cases: while keeping the signs of feedbacks the same as feedforward ones, the magnitudes of feedbacks are either randomized or set to uniform. The results are shown in Table \ref{tab:main} : Plus sign (+) denotes combination of methods. For example, uSF+BM means Batch Manhattan with uniform sign-concordant feedback. \textbf{SGD:} Stochastic gradient descent, the baseline algorithm. \textbf{BM:} SGD + Batch Manhattan. \textbf{BN:} SGD + Batch Normalization. \textbf{BN+BM:} SGD + Batch Normalization + Batch Manhattan. \textbf{uSF:} Uniform sign-concordant feedback. This condition often had exploding gradients.  \textbf{NuSF:} same as uSF, but with feedback weights normalized by dividing the number of inputs of the feedforward filter (filter width * filter height * input feature number). This scheme avoids the exploding gradients but still suffers from vanishing gradients. \textbf{uSF+BM:} this setting is somewhat unstable for small batch sizes. Two training procedures were explored: (1) batch size 100 for all epochs (2) batch size 100 for 3 epochs and then batch size 500 for the remaining epochs. The best performance was reported. While this gives a little advantage to this model since more settings were tried, we believe it is informative to isolate the stability issue and show what can be achieved if the model converges well. Note that uSF+BM is the only entry with slightly different training procedures. All other models share exactly the same training procedure \& parameters. \textbf{uSF+BN, uSF+BN+BM,  brSF+BN+BM, frSF+BN+BM:} These are some combinations of uSF, brSF, frSF, BN and BM. The last three are the most robust, well-performing ,and biologically-plausible algorithms. \\

\begin{table*}
\small
\centering
\begin{tabular}{|c|c|ccccc|ccccc|}
\hline
 &  Baseline &   \multicolumn{5} {c|} {\specialcell{ Part 1: Random $M$ and $S$ every batch} } &   \multicolumn{5} {c|} {\specialcell{  Part 2:  Fixed random $M$ and $S$} } \\ 
\hline
 Experiment B &SGD &100\% &75\% &50\% &25\% & \specialcell{Same Sign\\(brSF\\+BN+BM)} &100\% &75\% &50\% &25\% & \specialcell{Same Sign\\(frSF\\+BN+BM)}\\
\hline
MNIST &0.67 &\tdR{7.87} &\tdR{8.34} &\tdR{7.54} &1.01 &0.80 &\tdR{4.25} &3.56 &1.37 &0.84 &0.91\\
\rowcolor{LightCyan}
CIFAR &22.73 &\tdR{71.18} &\tdR{75.87} &\tdR{70.60} &19.98 &\tdG{18.44} &\tdR{71.41} &\tdR{68.49} &\tdR{31.54} &20.85 &\tdG{19.02}\\
CIFAR100 &55.15 &\tdR{93.72} &\tdR{96.23} &\tdR{94.58} &\tdR{68.98} &\tdG{50.74} &\tdR{96.26} &\tdR{95.22} &\tdR{71.98} &56.02 &52.25\\
\rowcolor{LightCyan}
SVHN &9.06 &\tdR{75.25} &\tdR{77.91} &\tdR{74.64} &10.94 &9.95 &\tdR{37.78} &\tdR{32.72} &\tdR{15.02} &11.50 &10.16\\
STL10 &48.01 &\tdR{69.65} &\tdR{72.50} &\tdR{72.10} &\tdG{44.82} &\tdG{42.74} &\tdR{72.46} &\tdR{68.96} &50.54 &\tdG{43.85} &\tdG{42.68}\\
\rowcolor{LightCyan}
Cal101 &74.08 &\tdR{91.46} &\tdR{93.99} &\tdR{91.36} &\tdG{67.65} &\tdG{59.54} &\tdR{94.20} &\tdR{87.57} &\tdG{68.60} &\tdG{63.12} &\tdG{60.27}\\
Cal256-101 &87.06 &\tdR{93.24} &\tdR{94.02} &\tdR{92.75} &\tdG{82.75} &\tdG{78.92} &\tdR{95.98} &\tdR{90.88} &84.22 &\tdG{83.04} &\tdG{80.59}\\
\rowcolor{LightCyan}
iCub &57.62 &\tdR{75.56} &\tdR{79.36} &\tdR{83.61} &\tdG{52.42} &\tdG{47.33} &\tdR{68.97} &\tdR{63.82} &\tdR{64.62} &\tdG{51.22} &\tdG{46.08}\\
Flowers17 &35.29 &\tdR{70.29} &\tdR{82.06} &\tdR{73.82} &35.00 &\tdG{29.41} &\tdR{79.12} &\tdR{61.76} &\tdR{44.12} &35.59 &\tdG{31.47}\\
\rowcolor{LightCyan}
Flowers102 &77.30 &\tdR{90.54} &\tdR{92.58} &\tdR{89.56} &\tdG{73.18} &\tdG{73.31} &\tdR{93.69} &\tdR{93.77} &77.62 &77.85 &\tdG{73.57}\\
PubFig83-ID &63.25 &\tdR{95.33} &\tdR{95.67} &\tdR{94.17} &61.25 &\tdG{42.67} &\tdR{94.25} &\tdR{89.25} &64.33 &\tdG{47.17} &\tdG{40.33}\\
\rowcolor{LightCyan}
SUFR-W-ID &80.00 &\tdR{95.00} &\tdR{95.75} &\tdR{92.50} &\tdG{71.00} &\tdG{66.25} &\tdR{95.50} &\tdR{95.50} &77.25 &\tdG{68.00} &\tdG{66.50}\\
LFW-ID &79.25 &\tdR{95.25} &\tdR{96.00} &\tdR{94.50} &\tdG{64.50} &\tdG{57.25} &\tdR{95.75} &\tdR{95.25} &\tdG{75.00} &\tdG{59.75} &\tdG{55.75}\\
\rowcolor{LightCyan}
Scene67 &87.16 &\tdR{94.48} &\tdR{94.78} &\tdR{93.43} &\tdG{83.13} &\tdG{82.31} &\tdR{95.37} &\tdR{92.69} &88.36 &84.40 &\tdG{81.79}\\
TIMIT80 &23.04 &\tdR{55.32} &\tdR{61.36} &\tdR{55.28} &\tdR{26.48} &25.24 &\tdR{62.60} &\tdR{33.48} &\tdR{27.56} &\tdR{26.08} &24.92\\
\hline
\end{tabular}
\caption { \textbf{Experiment B Part 1 (left):} Feedbacks have random magnitudes, varing probability of having different signs (percentages in second row, column 3-7) from the feedforward ones. The $M$ and $S$ redrawn in each mini-batch.  Numbers are error rates (\%).  \tdR{Yellow}: performances worse than baseline(SGD) by 3\% or more.  \tdG{Blue}: performances better than baseline(SGD) by 3\% or more.  \textbf{Experiment B Part 2 (right):} Same as part 1, but The $M$ and $S$ were fixed throughout each experiment.  } \label{tab:percent_diff}
\end{table*}

\newcolumntype{H}{>{\setbox0=\hbox\bgroup}c<{\egroup}@{}}
\begin{table*}
\centering
\small
\begin{tabular}{|c|Hcccccccc|}
\hline
Experiment C1 &SGD &SGD &RndF &NuSF &BN &\specialcell{RndF+BN} &\specialcell{RndF+BM} &\specialcell{RndF+BN+BM} &\specialcell{uSF+BN+BM}\\
\hline
MNIST &0.67 &0.67 &1.81 &0.60 &0.52 &0.83 &1.89 &1.07 &0.83\\
\rowcolor{LightCyan}
CIFAR &22.73 &22.73 &\tdR{42.69} &\tdR{40.60} &\tdG{16.75} &24.35 &\tdR{62.71} &\tdR{25.75} &\tdG{19.29}\\
CIFAR100 &55.15 &55.15 &\tdR{90.88} &\tdR{71.51} &\tdG{49.44} &\tdR{60.83} &\tdR{97.11} &\tdR{64.69} &53.12\\
\rowcolor{LightCyan}
SVHN &9.06 &9.06 &\tdR{12.35} &\tdR{14.55} &7.50 &\tdR{12.63} &\tdR{13.63} &\tdR{12.79} &9.67\\
STL10 &48.01 &48.01 &\tdR{57.80} &\tdR{56.53} &45.19 &\tdR{51.60} &\tdR{80.39} &47.39 &\tdG{41.55}\\
\rowcolor{LightCyan}
Cal101 &74.08 &74.08 &\tdR{88.51} &\tdG{70.50} &\tdG{66.07} &72.81 &\tdR{98.42} &\tdG{67.12} &\tdG{60.70}\\
Cal256-101 &87.06 &87.06 &\tdR{94.12} &85.98 &\tdG{82.94} &85.49 &\tdR{98.14} &\tdG{83.63} &\tdG{80.78}\\
\rowcolor{LightCyan}
iCub &57.62 &57.62 &\tdR{67.87} &\tdR{66.57} &\tdG{46.43} &58.82 &\tdR{84.41} &59.02 &\tdG{48.38}\\
Flowers17 &35.29 &35.29 &\tdR{69.41} &\tdR{42.65} &36.76 &\tdR{43.53} &\tdR{91.18} &38.24 &32.65\\
\rowcolor{LightCyan}
Flowers102 &77.30 &77.30 &\tdR{92.31} &77.92 &75.78 &\tdR{81.22} &\tdR{96.13} &78.99 &\tdG{73.20}\\
PubFig83-ID &63.25 &63.25 &\tdR{95.42} &\tdR{78.58} &\tdG{51.08} &\tdR{67.00} &\tdR{97.67} &\tdG{55.25} &\tdG{40.67}\\
\rowcolor{LightCyan}
SUFR-W-ID &80.00 &80.00 &\tdR{94.75} &\tdR{83.50} &\tdG{75.00} &77.75 &\tdR{97.25} &\tdG{69.00} &\tdG{65.75}\\
LFW-ID &79.25 &79.25 &\tdR{95.75} &\tdR{85.75} &\tdG{73.75} &79.25 &\tdR{97.75} &\tdG{67.50} &\tdG{56.25}\\
\rowcolor{LightCyan}
Scene67 &87.16 &87.16 &\tdR{95.75} &88.21 &86.04 &87.84 &\tdR{97.69} &87.09 &\tdG{81.87}\\
TIMIT80 &23.04 &23.04 &\tdR{26.76} &\tdR{29.28} &23.92 &\tdR{26.52} &\tdR{33.12} &\tdR{26.32} &25.12\\
\hline
\hline
Experiment C2 &SGD &\specialcell{SGD\\ \textit{Bottom}} &\specialcell{SGD\\ \textit{Top}} &\specialcell{RndF\\ \textit{Bottom}} &\specialcell{RndF\\ \textit{Top}} &\specialcell{RndF+BN+BM\\ \textit{Bottom}} &\specialcell{RndF+BN+BM\\ \textit{Top}} &\specialcell{uSF+BN+BM\\ \textit{Bottom}} &\specialcell{uSF+BN+BM\\ \textit{Top}}\\
\hline
MNIST &0.67 &0.65 &\tdR{3.85} &\tdR{85.50} &\tdR{3.81} &\tdR{86.74} &\tdR{3.81} &0.66 &\tdR{3.85}\\
\rowcolor{LightCyan}
CIFAR &22.73 &23.12 &\tdR{56.80} &\tdR{89.71} &\tdR{56.77} &\tdR{78.90} &\tdR{58.54} &\tdG{16.72} &\tdR{57.84}\\
CIFAR100 &55.15 &\tdR{59.49} &\tdR{80.71} &\tdR{98.79} &\tdR{80.65} &\tdR{98.69} &\tdR{84.34} &\tdR{61.61} &\tdR{84.10}\\
\rowcolor{LightCyan}
SVHN &9.06 &8.31 &\tdR{75.22} &\tdR{82.86} &\tdR{75.12} &\tdR{84.84} &\tdR{69.99} &10.96 &\tdR{71.89}\\
STL10 &48.01 &49.96 &\tdR{74.69} &\tdR{88.36} &\tdR{72.44} &\tdR{81.31} &\tdR{76.11} &\tdR{52.18} &\tdR{76.09}\\
\rowcolor{LightCyan}
Cal101 &74.08 &71.97 &\tdR{82.72} &\tdR{98.63} &\tdR{79.14} &\tdR{98.21} &\tdR{80.40} &\tdG{63.86} &\tdR{79.98}\\
Cal256-101 &87.06 &86.08 &89.71 &\tdR{98.43} &88.92 &\tdR{98.14} &89.02 &\tdG{82.84} &89.12\\
\rowcolor{LightCyan}
iCub &57.62 &\tdG{46.73} &\tdR{83.96} &\tdR{87.56} &\tdR{83.26} &\tdR{80.36} &\tdR{84.31} &\tdG{49.33} &\tdR{84.16}\\
Flowers17 &35.29 &38.24 &\tdR{70.59} &\tdR{92.35} &\tdR{70.00} &\tdR{87.35} &\tdR{77.06} &\tdR{45.00} &\tdR{77.06}\\
\rowcolor{LightCyan}
Flowers102 &77.30 &78.99 &\tdR{86.57} &\tdR{97.89} &\tdR{86.84} &\tdR{98.11} &\tdR{84.24} &78.09 &\tdR{84.57}\\
PubFig83-ID &63.25 &\tdR{66.75} &\tdR{89.58} &\tdR{97.67} &\tdR{89.58} &\tdR{97.67} &\tdR{89.67} &\tdG{43.83} &\tdR{89.50}\\
\rowcolor{LightCyan}
SUFR-W-ID &80.00 &80.50 &\tdR{90.50} &\tdR{97.50} &\tdR{90.50} &\tdR{97.50} &\tdR{89.50} &\tdG{71.50} &\tdR{89.50}\\
LFW-ID &79.25 &79.75 &\tdR{92.50} &\tdR{98.25} &\tdR{93.00} &\tdR{97.00} &\tdR{89.50} &\tdG{65.00} &\tdR{89.50}\\
\rowcolor{LightCyan}
Scene67 &87.16 &88.73 &\tdR{91.57} &\tdR{97.84} &\tdR{91.49} &\tdR{97.54} &\tdR{91.19} &85.97 &\tdR{91.04}\\
TIMIT80 &23.04 &23.52 &\tdR{46.20} &\tdR{95.00} &\tdR{46.20} &\tdR{93.00} &\tdR{39.76} &24.96 &\tdR{40.16}\\
\hline
\end{tabular}
\caption { \textbf{Experiment C1:} fixed random feedbacks.  \textbf{Experiment C2:} (.)$\textit{Bottom}$: The model's last layer is initialized randomly and clamped/frozen. All learning happens in the layers before the last layer. (.)$\textit{Top}$: The model's layers before the last layer are initialized randomly and clamped/frozen. All learning happens in the last layer. Numbers are error rates (\%).  \tdR{Yellow}: performances worse than baseline(SGD) by 3\% or more.  \tdG{Blue}: performances better than baseline(SGD) by 3\% or more.  } \label{tab:exp_C_and_D}
\end{table*}

\begin{table*}
\centering
\begin{tabular}{|c|cccccccc|}
\hline
 &SGD &BM1 &BM2 &BM3 &\specialcell{uSF+BN} &\specialcell{uSF+BN+BM1} &\specialcell{uSF+BN+BM2} &\specialcell{uSF+BN+BM3}\\
\hline
MNIST &0.67 &0.99 &1.30 &1.09 &0.55 &0.83 &0.72 &0.61\\
\rowcolor{LightCyan}
CIFAR &22.73 &23.98 &23.09 &20.47 &\tdG{19.48} &\tdG{19.29} &\tdG{18.87} &\tdG{18.38}\\
CIFAR100 &55.15 &\tdR{58.44} &\tdR{58.81} &52.82 &57.19 &53.12 &52.38 &54.68\\
\rowcolor{LightCyan}
SVHN &9.06 &10.77 &\tdR{12.31} &\tdR{12.23} &8.73 &9.67 &10.54 &9.20\\
STL10 &48.01 &\tdG{44.14} &\tdG{44.74} &45.23 &48.49 &\tdG{41.55} &47.71 &46.45\\
\rowcolor{LightCyan}
Cal101 &74.08 &\tdG{66.70} &\tdG{65.96} &\tdG{70.28} &\tdG{63.33} &\tdG{60.70} &\tdG{64.38} &\tdG{62.59}\\
\hline
\end{tabular}
\caption {Different settings of Batch Manhattan (as described in Section \ref{sec:normalizations}) seem to give similar performances. SGD: setting 0, BM1: setting 1, BM2: setting 2, BM3: setting 3. The interaction of BM with sign concordant feedback weights (uSF) and Batch Normalization are shown in ``uSF+BN+(.)'' entries.  Numbers are error rates (\%). \tdR{Yellow}: performances worse than baseline (SGD) by 3\% or more.  \tdG{Blue}: performances better than baseline(SGD) by 3\% or more. } \label{tab:3BM}
\end{table*}

\subsubsection{Experiment B: Violating Sign-Concordance with probability p} 

In this experiment, we test  the effect of \textbf{partial sign-concordance}. That is, we test settings 4 and 5 as described in Section \ref{sec:abp}.  In these cases, the feedback weight magnitudes were all random. Strict sign-concordance was relaxed by manipulating the probability $p$ of concordance between feedforward and feedback weight signs. Feedback weights $V = M \circ sign(W) \circ S_p$ depend on the matrix $S_p$ as defined in Section \ref{sec:abp}. Table \ref{tab:percent_diff} Part 1 reports results from setting 4, the case where a new $M$ and $S_p$ is sampled for each batch. Table \ref{tab:percent_diff} Part 2 reports results of setting 5, the case where $M$ and $S_p$ are fixed. The main observation from this experiment is that the performance declines as the level of sign-concordance decreases.

\subsubsection{Experiment C1: Fixed Random Feedback } 

 Results are shown in Table \ref{tab:exp_C_and_D}: \textbf{RndF:} fixed random feedbacks. \textbf{RndF+BN, RndF+BN+BM:} some combinations of RndF, BN and BM. \textbf{uSF+BN+BM:} one of our best algorithms, for reference. The ``RndF'' setting is the same as the one proposed by \cite{lillicrap2014random}. Apparently it does not perform well on most datasets. However, combining it with Batch Normalization makes it significantly better. Although it remains  somewhat worse than its sign concordant counterpart. Another observation is that random feedback does not work well with BM alone (but better with BN+BM).

\subsubsection{Experiment C2: Control experiments for Experiment C1} 

 The fact that random feedback weights can learn meaningful representations is somewhat surprising. We explore this phenomenon by running some control experiments, where we run two control models for each model of interest: \textbf{1.} (.)\textit{Bottom}: The model's last layer is initialized randomly and clamped/frozen. All learning happens in the layers before the last layer.  \textbf{2.} (.)\textit{Top}: The model's layers before the last layer are initialized randomly and clamped/frozen. All learning happens in the last layer. Results are shown in Table \ref{tab:exp_C_and_D}.

There are some observations: \textbf{(i)} When only the last layer was allowed to adapt, all models behaved similarly. This was expected since the models only differed in the way they backpropagate errors. \textbf{(ii)} With the last layer is clamped, random feedback cannot learn meaningful representations. \textbf{(iii)} In contrast, the models with sign concordant feedback can learn surprisingly good representations even with the last layer locked. We can draw the following conclusions from such observations: sign concordant feedback  ensures that meaningful error signals reach lower layers by itself, while random feedback is not sufficient. If all layers can learn, random feedback seems to work via a ``mysterious co-adaptation'' between the last layer and the layers before it. This ``mysterious co-adaptation'' was first observed by \cite{lillicrap2014random}, where it was called ``feedback alignment'' and some analyses were given. Note that our Experiment C shows that the effect is more significant if Batch Normalization is applied.

\subsubsection{Miscellaneous Experiment: different settings of Batch Manhattan} 

We show that the 3 settings of BM (as described in Section \ref{sec:normalizations}) produce similar performances. This is the case for both symmetric and asymmetric BPs. Results are in Table \ref{tab:3BM}.

\section{Discussion}
This work aims to establish that there exist variants of the gradient backpropagation algorithm that could plausibly be implemented in the brain. To that end we considered the question: how important is weight symmetry to backpropagation? Through a series of experiments with increasingly  asymmetric backpropagation algorithms, our work complements a recent  demonstration\cite{lillicrap2014random}  that perfect weight symmetry can be significantly relaxed while still retaining strong performance. 

These results show that Batch Normalization and/or Batch Manhattan are crucial for asymmetric backpropagation to work. Furthermore, they are complementary operations that are better used together than individually. These results highlight the importance of sign-concordance to asymmetric backpropagation by systematically exploring how performance declines with its relaxation.

Finally, let us return to our original motivation. How does all this relate to the brain? Based on current neuroscientific understanding of cortical feedback, we cannot make any claim about whether such asymmetric BP algorithms are actually implemented by the brain. Nevertheless, this work shows that asymmetric BPs, while having less constraints, are not computationally inferior to standard BP. So if the brain were to implement one of them, it could obtain most of the benefits of the standard algorithm. 

This work suggests a hypothesis that could be checked by empirical neuroscience research: if the brain does indeed implement an asymmetric BP algorithm, then there is likely to be a high degree of sign-concordance in cortical forward-backward connections.

These empirical observations concerning Batch Manhattan updating may shed light on the general issue of how synaptic plasticity may implement learning algorithms. They show that changes of synaptic strength can be rather noisy. That is, the \textit{sign} of a long term accumulation of synaptic potentiation or depression, rather than precise magnitude, is what is important. This scheme seems biologically implementable.

\section{Acknowledgements}
We thank G. Hinton for useful comments. This work was supported  by the Center for Brains, Minds and Machines (CBMM), funded by NSF STC award  CCF – 1231216.

%\fontsize{9.5pt}{10.5pt} \selectfont

\bibliographystyle{aaai}

\begin{thebibliography}{}

\end{thebibliography}


\begin{thebibliography}{}

\bibitem[\protect\citeauthoryear{Abdel-Hamid \bgroup et al\mbox.\egroup
  }{2012}]{Abdel-Hamid2012}
Abdel-Hamid, O.; Mohamed, A.; Jiang, H.; and Penn, G.
\newblock 2012.
\newblock {Applying convolutional neural networks concepts to hybrid NN-HMM
  model for speech recognition}.
\newblock In {\em IEEE International Conference on Acoustics, Speech and Signal
  Processing (ICASSP)},  4277--4280.

\bibitem[\protect\citeauthoryear{Bengio \bgroup et al\mbox.\egroup
  }{2015}]{bengio2015towards}
Bengio, Y.; Lee, D.-H.; Bornschein, J.; and Lin, Z.
\newblock 2015.
\newblock Towards biologically plausible deep learning.
\newblock {\em arXiv preprint arXiv:1502.04156}.

\bibitem[\protect\citeauthoryear{Bengio}{2014}]{bengio2014auto}
Bengio, Y.
\newblock 2014.
\newblock How auto-encoders could provide credit assignment in deep networks
  via target propagation.
\newblock {\em arXiv preprint arXiv:1407.7906}.

\bibitem[\protect\citeauthoryear{Chinta and Tweed}{2012}]{chinta2012adaptive}
Chinta, L.~V., and Tweed, D.~B.
\newblock 2012.
\newblock Adaptive optimal control without weight transport.
\newblock {\em Neural computation} 24(6):1487--1518.

\bibitem[\protect\citeauthoryear{Coates, Ng, and
  Lee}{2011}]{coates2011analysis}
Coates, A.; Ng, A.~Y.; and Lee, H.
\newblock 2011.
\newblock An analysis of single-layer networks in unsupervised feature
  learning.
\newblock In {\em International conference on artificial intelligence and
  statistics},  215--223.

\bibitem[\protect\citeauthoryear{Crick}{1989}]{crick1989recent}
Crick, F.
\newblock 1989.
\newblock The recent excitement about neural networks.
\newblock {\em Nature} 337(6203):129--132.

\bibitem[\protect\citeauthoryear{Fanello \bgroup et al\mbox.\egroup
  }{2013}]{fanello2013icub}
Fanello, S.~R.; Ciliberto, C.; Santoro, M.; Natale, L.; Metta, G.; Rosasco, L.;
  and Odone, F.
\newblock 2013.
\newblock icub world: Friendly robots help building good vision data-sets.
\newblock In {\em Computer Vision and Pattern Recognition Workshops (CVPRW),
  2013 IEEE Conference on},  700--705.
\newblock IEEE.

\bibitem[\protect\citeauthoryear{Fei-Fei, Fergus, and
  Perona}{2007}]{fei2007learning}
Fei-Fei, L.; Fergus, R.; and Perona, P.
\newblock 2007.
\newblock Learning generative visual models from few training examples: An
  incremental bayesian approach tested on 101 object categories.
\newblock {\em Computer Vision and Image Understanding} 106(1):59--70.

\bibitem[\protect\citeauthoryear{Garofolo \bgroup et al\mbox.\egroup
  }{}]{TIMITdatabase}
Garofolo, J.; Lamel, L.; Fisher, W.; Fiscus, J.; Pallett, D.; Dahlgren, N.; and
  Zue, V.
\newblock Timit acoustic-phonetic continuous speech corpus.

\bibitem[\protect\citeauthoryear{Graves, Wayne, and
  Danihelka}{2014}]{graves2014neural}
Graves, A.; Wayne, G.; and Danihelka, I.
\newblock 2014.
\newblock Neural turing machines.
\newblock {\em arXiv preprint arXiv:1410.5401}.

\bibitem[\protect\citeauthoryear{Griffin, Holub, and
  Perona}{2007}]{griffin2007caltech}
Griffin, G.; Holub, A.; and Perona, P.
\newblock 2007.
\newblock Caltech-256 object category dataset.

\bibitem[\protect\citeauthoryear{Grossberg}{1987}]{grossberg1987competitive}
Grossberg, S.
\newblock 1987.
\newblock Competitive learning: From interactive activation to adaptive
  resonance.
\newblock {\em Cognitive science} 11(1):23--63.

\bibitem[\protect\citeauthoryear{Hinton and
  McClelland}{1988}]{hinton1988learning}
Hinton, G.~E., and McClelland, J.~L.
\newblock 1988.
\newblock Learning representations by recirculation.
\newblock In {\em Neural information processing systems},  358--366.
\newblock New York: American Institute of Physics.

\bibitem[\protect\citeauthoryear{Hinton and
  Salakhutdinov}{2006}]{hinton2006reducing}
Hinton, G.~E., and Salakhutdinov, R.~R.
\newblock 2006.
\newblock Reducing the dimensionality of data with neural networks.
\newblock {\em Science} 313(5786):504--507.

\bibitem[\protect\citeauthoryear{Hinton \bgroup et al\mbox.\egroup
  }{2012}]{hinton2012deep}
Hinton, G.; Deng, L.; Yu, D.; Dahl, G.~E.; Mohamed, A.-r.; Jaitly, N.; Senior,
  A.; Vanhoucke, V.; Nguyen, P.; Sainath, T.~N.; et~al.
\newblock 2012.
\newblock Deep neural networks for acoustic modeling in speech recognition: The
  shared views of four research groups.
\newblock {\em Signal Processing Magazine, IEEE} 29(6):82--97.

\bibitem[\protect\citeauthoryear{Hornik, Stinchcombe, and
  White}{1989}]{hornik1989multilayer}
Hornik, K.; Stinchcombe, M.; and White, H.
\newblock 1989.
\newblock Multilayer feedforward networks are universal approximators.
\newblock {\em Neural networks} 2(5):359--366.

\bibitem[\protect\citeauthoryear{Huang \bgroup et al\mbox.\egroup
  }{2008}]{Huang2008}
Huang, G.~B.; Mattar, M.; Berg, T.; and Learned-Miller, E.
\newblock 2008.
\newblock {Labeled faces in the wild: A database for studying face recognition
  in unconstrained environments}.
\newblock In {\em Workshop on faces in real-life images: Detection, alignment
  and recognition (ECCV)}.

\bibitem[\protect\citeauthoryear{Ioffe and Szegedy}{2015}]{ioffe2015batch}
Ioffe, S., and Szegedy, C.
\newblock 2015.
\newblock Batch normalization: Accelerating deep network training by reducing
  internal covariate shift.
\newblock {\em arXiv preprint arXiv:1502.03167}.

\bibitem[\protect\citeauthoryear{Krizhevsky, Sutskever, and
  Hinton}{2012}]{Krizhevsky2012}
Krizhevsky, A.; Sutskever, I.; and Hinton, G.
\newblock 2012.
\newblock {ImageNet classification with deep convolutional neural networks}.
\newblock In {\em Advances in neural information processing systems}.

\bibitem[\protect\citeauthoryear{Krizhevsky}{2009}]{krizhevsky2009learning}
Krizhevsky, A.
\newblock 2009.
\newblock Learning multiple layers of features from tiny images.

\bibitem[\protect\citeauthoryear{Le~Cun}{1986}]{le1986learning}
Le~Cun, Y.
\newblock 1986.
\newblock Learning process in an asymmetric threshold network.
\newblock In {\em Disordered systems and biological organization}. Springer.
\newblock  233--240.

\bibitem[\protect\citeauthoryear{LeCun, Cortes, and Burges}{}]{mnistWebsite}
LeCun, Y.; Cortes, C.; and Burges, C.~J.
\newblock The mnist database.

\bibitem[\protect\citeauthoryear{Leibo, Liao, and
  Poggio}{2014}]{leibo2014subtasks}
Leibo, J.~Z.; Liao, Q.; and Poggio, T.
\newblock 2014.
\newblock {Subtasks of Unconstrained Face Recognition}.
\newblock In {\em International Joint Conference on Computer Vision, Imaging
  and Computer Graphics, VISIGRAPP}.

\bibitem[\protect\citeauthoryear{Lillicrap \bgroup et al\mbox.\egroup
  }{2014}]{lillicrap2014random}
Lillicrap, T.~P.; Cownden, D.; Tweed, D.~B.; and Akerman, C.~J.
\newblock 2014.
\newblock Random feedback weights support learning in deep neural networks.
\newblock {\em arXiv preprint arXiv:1411.0247}.

\bibitem[\protect\citeauthoryear{Mazzoni, Andersen, and
  Jordan}{1991}]{mazzoni1991more}
Mazzoni, P.; Andersen, R.~A.; and Jordan, M.~I.
\newblock 1991.
\newblock A more biologically plausible learning rule for neural networks.
\newblock {\em Proceedings of the National Academy of Sciences}
  88(10):4433--4437.

\bibitem[\protect\citeauthoryear{Mikolov \bgroup et al\mbox.\egroup
  }{2013}]{mikolov2013distributed}
Mikolov, T.; Sutskever, I.; Chen, K.; Corrado, G.~S.; and Dean, J.
\newblock 2013.
\newblock Distributed representations of words and phrases and their
  compositionality.
\newblock In {\em Advances in Neural Information Processing Systems (NIPS)},
  3111--3119.

\bibitem[\protect\citeauthoryear{Netzer \bgroup et al\mbox.\egroup
  }{2011}]{netzer2011reading}
Netzer, Y.; Wang, T.; Coates, A.; Bissacco, A.; Wu, B.; and Ng, A.~Y.
\newblock 2011.
\newblock Reading digits in natural images with unsupervised feature learning.
\newblock In {\em NIPS workshop on deep learning and unsupervised feature
  learning}, volume 2011, ~5.

\bibitem[\protect\citeauthoryear{Nilsback and
  Zisserman}{2006}]{nilsback2006visual}
Nilsback, M.-E., and Zisserman, A.
\newblock 2006.
\newblock A visual vocabulary for flower classification.
\newblock In {\em Computer Vision and Pattern Recognition, 2006 IEEE Computer
  Society Conference on}.
\newblock IEEE.

\bibitem[\protect\citeauthoryear{Nilsback and
  Zisserman}{2008}]{nilsback2008automated}
Nilsback, M.-E., and Zisserman, A.
\newblock 2008.
\newblock Automated flower classification over a large number of classes.
\newblock In {\em Computer Vision, Graphics \& Image Processing, 2008.
  ICVGIP'08. Sixth Indian Conference on}.
\newblock IEEE.

\bibitem[\protect\citeauthoryear{O'Reilly}{1996}]{o1996biologically}
O'Reilly, R.~C.
\newblock 1996.
\newblock Biologically plausible error-driven learning using local activation
  differences: The generalized recirculation algorithm.
\newblock {\em Neural computation} 8(5):895--938.

\bibitem[\protect\citeauthoryear{Pinto \bgroup et al\mbox.\egroup
  }{2011}]{pinto2011scaling}
Pinto, N.; Stone, Z.; Zickler, T.; and Cox, D.
\newblock 2011.
\newblock Scaling up biologically-inspired computer vision: A case study in
  unconstrained face recognition on facebook.
\newblock In {\em Computer Vision and Pattern Recognition Workshops (CVPRW),
  2011 IEEE Computer Society Conference on},  35--42.
\newblock IEEE.

\bibitem[\protect\citeauthoryear{Quattoni and
  Torralba}{2009}]{quattoni2009recognizing}
Quattoni, A., and Torralba, A.
\newblock 2009.
\newblock Recognizing indoor scenes.
\newblock In {\em Computer Vision and Pattern Recognition, 2009. CVPR 2009.
  IEEE Conference on},  413--420.
\newblock IEEE.

\bibitem[\protect\citeauthoryear{Riedmiller and
  Braun}{1993}]{riedmiller1993direct}
Riedmiller, M., and Braun, H.
\newblock 1993.
\newblock A direct adaptive method for faster backpropagation learning: The
  rprop algorithm.
\newblock In {\em Neural Networks, 1993., IEEE International Conference on},
  586--591.
\newblock IEEE.

\bibitem[\protect\citeauthoryear{Rumelhart, Hinton, and
  Williams}{1988}]{rumelhart1988learning}
Rumelhart, D.~E.; Hinton, G.~E.; and Williams, R.~J.
\newblock 1988.
\newblock Learning representations by back-propagating errors.
\newblock {\em Cognitive modeling}.

\bibitem[\protect\citeauthoryear{Smolensky}{1986}]{smolensky1986information}
Smolensky, P.
\newblock 1986.
\newblock Information processing in dynamical systems: Foundations of harmony
  theory.

\bibitem[\protect\citeauthoryear{Stellwagen and
  Malenka}{2006}]{stellwagen2006synaptic}
Stellwagen, D., and Malenka, R.~C.
\newblock 2006.
\newblock Synaptic scaling mediated by glial tnf-$\alpha$.
\newblock {\em Nature} 440(7087):1054--1059.

\bibitem[\protect\citeauthoryear{Taigman \bgroup et al\mbox.\egroup
  }{2014}]{taigman2014deepface}
Taigman, Y.; Yang, M.; Ranzato, M.; and Wolf, L.
\newblock 2014.
\newblock Deepface: Closing the gap to human-level performance in face
  verification.
\newblock In {\em Computer Vision and Pattern Recognition (CVPR), 2014 IEEE
  Conference on},  1701--1708.
\newblock IEEE.

\bibitem[\protect\citeauthoryear{Turrigiano and Nelson}{2004}]{Turrigiano2004}
Turrigiano, G.~G., and Nelson, S.~B.
\newblock 2004.
\newblock {Homeostatic plasticity in the developing nervous system}.
\newblock {\em Nature Reviews Neuroscience}.

\bibitem[\protect\citeauthoryear{Turrigiano}{2008}]{turrigiano2008self}
Turrigiano, G.~G.
\newblock 2008.
\newblock The self-tuning neuron: synaptic scaling of excitatory synapses.
\newblock {\em Cell} 135(3):422--435.

\bibitem[\protect\citeauthoryear{Vedaldi and Lenc}{2015}]{vedaldi15matconvnet}
Vedaldi, A., and Lenc, K.
\newblock 2015.
\newblock MatConvNet -- Convolutional Neural Networks for MATLAB
%\newblock In {\em Proceeding of the {ACM} Int. Conf. on Multimedia}.

\bibitem[\protect\citeauthoryear{Zamanidoost \bgroup et al\mbox.\egroup
  }{2015}]{zamanidoostmanhattan}
Zamanidoost, E.; Bayat, F.~M.; Strukov, D.; and Kataeva, I.
\newblock 2015.
\newblock Manhattan rule training for memristive crossbar circuit pattern
  classifiers.


\bibitem[\protect\citeauthoryear{Zamanidoost \bgroup et al\mbox.\egroup
  }{2015}]{zamanidoostmanhattan}
Zamanidoost, E.; Bayat, F.~M.; Strukov, D.; and Kataeva, I.
\newblock 2015.
\newblock Manhattan rule training for memristive crossbar circuit pattern
  classifiers.


\end{thebibliography}

\end{document}